%% file: PaperForReview.tex
\documentclass[10pt,twocolumn,letterpaper]{article}
\usepackage[accsupp]{axessibility}  % Improves PDF readability for those with disabilities.
\usepackage{wacv}              % To produce the CAMERA-READY version
\usepackage{graphicx}
\usepackage{amsmath}
\usepackage{amssymb}
\usepackage{booktabs}
\usepackage{times}
\usepackage{epsfig}
\usepackage{graphicx}
\usepackage{amsmath}
\usepackage{amssymb}
\usepackage{times}
\usepackage{microtype}
\usepackage{epsfig}
\usepackage[table,xcdraw]{xcolor}
\usepackage{caption}
\usepackage{float}
\usepackage{placeins}
\usepackage{color, colortbl}
\usepackage{stfloats}
\usepackage{enumitem}
\usepackage{tabularx}
\usepackage{xstring}
\usepackage{multirow}
\usepackage{xspace}
\usepackage{url}
\usepackage{subcaption}
\usepackage{xcolor}
\usepackage[hang,flushmargin]{footmisc}
\usepackage{makecell}
\usepackage{adjustbox}
\usepackage{booktabs}
\usepackage[pagebackref,breaklinks,colorlinks]{hyperref}

\usepackage[capitalize]{cleveref}
\crefname{section}{Sec.}{Secs.}
\Crefname{section}{Section}{Sections}
\Crefname{table}{Table}{Tables}
\crefname{table}{Tab.}{Tabs.}

\def\wacvPaperID{2151} % *** Enter the WACV Paper ID here
\def\confName{WACV}
\def\confYear{2024}

\begin{document}

\title{PMI Sampler: Patch Similarity Guided Frame Selection For Aerial Action Recognition}

\author{Ruiqi Xian, Xijun Wang, Divya Kothandaraman, Dinesh Manocha\\
University of Maryland - College Park\\
{\tt\small \{rxian, xijun, dkr, dmanocha\}@umd.edu}
% \and
}
\maketitle
% \footnote{An extended version can be found: \href{https://arxiv.org/abs/2304.06866}{https://arxiv.org/abs/2304.06866}}
%%%%%%%%% ABSTRACT
\input{00_abstract}

%%%%%%%%% BODY TEXT
\input{figs/Example}

\input{01_intro}
\input{figs/PMI_score}
\input{02_related}

\input{figs/Sampler}
\input{03_method}

\input{figs/ReLu}
\input{04_results}
\input{10_conclusion}

%%%%%%%%% REFERENCES
\clearpage
{\small
\bibliographystyle{ieee_fullname}
\bibliography{egbib}
}
\clearpage
\input{12_appendix}
\end{document}

%% file: 00_abstract.tex
\begin{abstract}
% Abstract goes here.
% \footnote{An extended version can be found: \href{https://arxiv.org/abs/2304.06866}{https://arxiv.org/abs/2304.06866}} 
% \footnote{An extended version can be found: https://arxiv.org/abs/2304.06866} 
We present a new algorithm for the selection of informative frames in video action recognition.
% that takes into account semantically rich content and temporal information redundancy. 
Our approach is designed for aerial videos captured using a moving camera where human actors occupy a small spatial resolution of video frames.  Our algorithm utilizes the motion bias within aerial videos, which enables the selection of motion-salient frames. %for training. 
We introduce the concept of patch mutual information (PMI) score to quantify the motion bias between adjacent frames, by measuring the similarity of patches. We use this score to assess the amount of discriminative motion information contained in one frame relative to another. We present an adaptive frame selection strategy using shifted leaky ReLu and cumulative distribution function, which ensures that the sampled frames %uniformly
%fully?
comprehensively cover all the essential segments with high motion salience.
% or abrupt viewpoint changes
Our approach can be integrated with any action recognition model to enhance its accuracy. In practice, our method achieves a relative improvement of 2.2 - 13.8\% in top-1 accuracy on UAV-Human, 6.8\% on NEC Drone, and 9.0\% on Diving48 datasets. The code is available at \href{https://github.com/Ricky-Xian/PMI-Sampler}{https://github.com/Ricky-Xian/PMI-Sampler}.
% \href{https://github.com/Ricky-Xian/PMI-Sampler}{https://github.com/Ricky-Xian/PMI-Sampler}.

\end{abstract}

%% file: figs/Example.tex
\begin{figure*}[tp]
    \centering
    % \vspace{-70mm}
    \includegraphics[width=\textwidth]{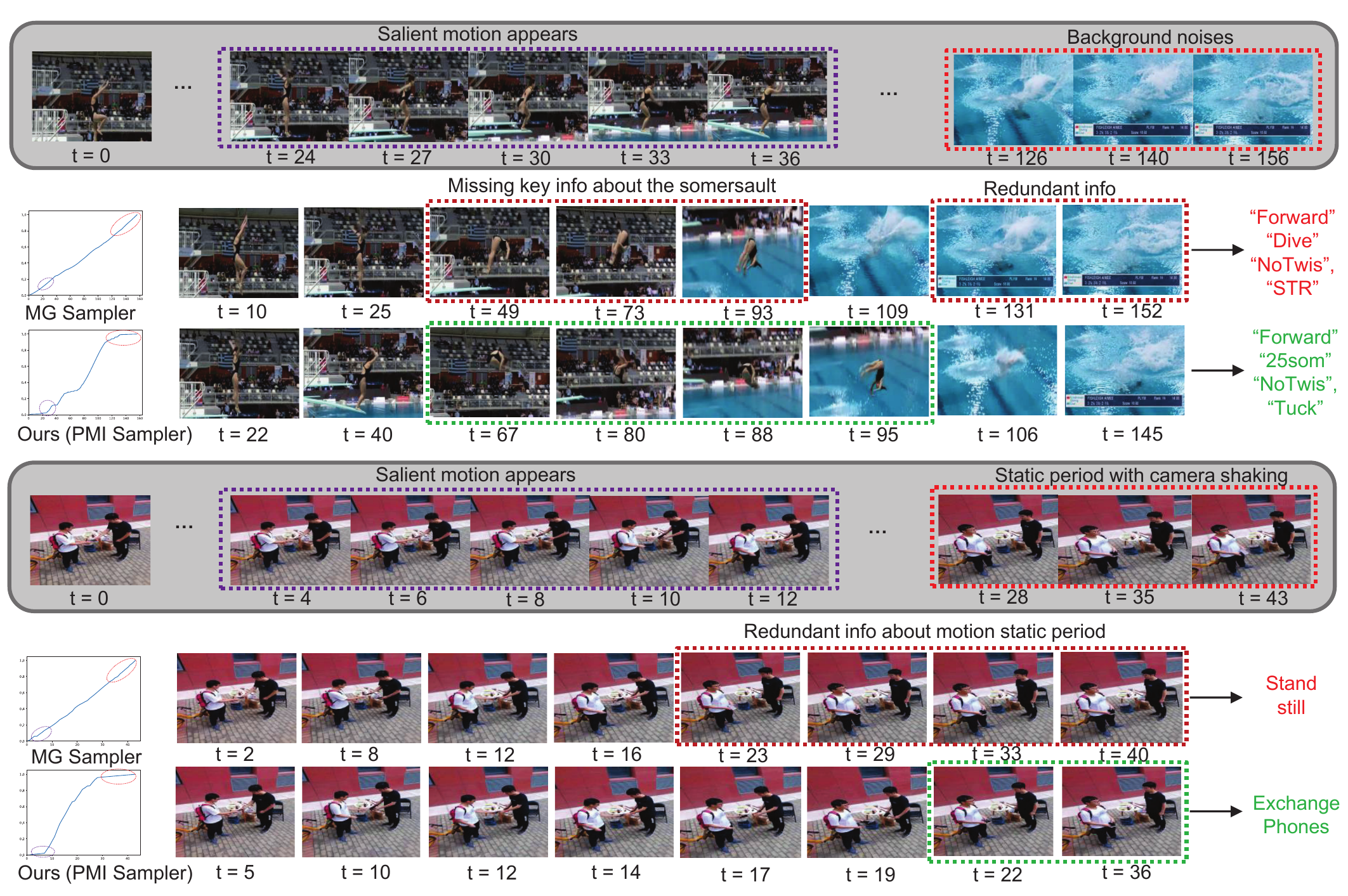}
    % \vspace{-70mm}
    % \includesvg[width=\linewidth]{figs/ReLu.svg}
    \caption{Sample eight frames from typical videos in Diving48 and UAV-Human. Compared with the state-of-the-art, MG Sampler~\cite{zhi2021mgsampler}, our approach can better represent the motion distribution and provides an easier way to distinguish motion salient frames in videos. PMI Sampler is more robust to noises and outliers and can better handle the background changes caused by the moving camera. }
    \label{fig:example}
\vspace{-4mm}
\end{figure*}

%% file: 01_intro.tex
\section{Introduction}

The applications of unmanned aerial vehicles (UAVs) have been expanding rapidly into search and rescue, agriculture, security, surveillance, etc. This is giving rise to many challenging problems including detection, re-identification, tracking, and recognition. Among these tasks, aerial video action recognition is regarded as one of the most difficult problems.
%due to the rich semantics and redundant information contained in video content. 
Despite the success of deep learning-based methods in video action recognition on ground camera footage~\cite{feichtenhofer2020x3d,kondratyuk2021movinets,Monfort2020MomentsIT}, current methods do not result in high accuracy on aerial videos. 
%data results in a significant accuracy drop. 
%This is attributed to the large domain shift between aerial and ground videos, which presents a formidable obstacle to researchers seeking to develop effective algorithms for aerial action recognition.

Aerial video data are often captured by a camera mounted on a moving UAV in oblique or overhead angles.
%which rarely exist in generic videos. 
The resulting footage features human actors that appear significantly smaller (typically less than 10\% pixels), due to the high camera altitude, with a large portion of the video frame dominated by the background information. The size and scale of the human actor may vary considerably, owing to changes in flying altitude during video data collection. Furthermore, the continuous movement of the UAV can cause the camera viewing angle shifting, resulting in motion blur and occlusion. These factors collectively make the development of accurate algorithms for aerial action recognition a challenging task.

% Because of high camera altitude and moving camera in aerial videos, many parts of the human body are not visible. Some parts of the human body that result in the action may be occluded by some other parts that do not contribute to the action. Also, there are lots of "duplicated" frames because of the high frame rate, which essentially contains the redundant information. Therefore, not all the video frames are useful for the training and using some of them may even decrease the overall accuracy. 

The decrease in performance of generic video recognition models on aerial data can be reduced through the use of motion-guided frame sampling during training~\cite{kothandaraman2022differentiable,mitfas2023}.
However, current deep learning-based action recognition methods mostly use fixed hand-crafted sampling techniques for video analysis~\cite{Simonyan2014TwoStreamCN,Tran2017ACL,Wang2016TemporalSN}. Typically, frames are randomly sampled in a uniform manner or successively with a fixed stride from the original video. This fixed sampling strategy can be sub-optimal for several reasons. First, the motion duration varies for different videos and actions, and fixed sampling may not capture the entire motion duration, potentially overlooking some useful information. Second, the sampling approach should prioritize discriminative frames over redundant or uninformative background frames, as not all frames are equally useful in terms of recognition.

% One possible solution to reduce the performance drop of generic video recognition model on aerial data is to sample frames that contain more motion information for training. However, over the last decade, deep learning based action recognition methods commonly use a fixed hand-crafted sampling technique for video understanding~\cite{Simonyan2014TwoStreamCN,Tran2017ACL,Wang2016TemporalSN}. Typically, frames are randomly sampled either uniformly or successively with a fixed stride from the original video. However, such fixed sampling strategy has a few drawbacks. First, duration of the motion varies for different videos and actions. If the sampling is always fixed across videos, it is likely that sampled frames may not cover the whole motion duration and key information is neglected. Second, sampling should focus more on discriminative frames rather than 'duplicated' frames or pointless background frames since not all the frames are equally important for recognition. 
% Based on the analysis above, further study is still required for a principled and adaptive sampling technique for UAV video action recognition.

Recently, some techniques~\cite{Wu2018AdaFrameAF,Korbar2019SCSamplerSS,ghodrati2021frameexit,lin2022ocsampler,gao2020listen} have been proposed for frame selection by modeling it as a decision making task. Typically, these methods employ a learning-based module to sequentially select more informative frames or to conditionally exit early. While these methods have shown promising results, their performance heavily relies on the training data and may not easily transfer to unseen domains. Unfortunately, the scarcity of labeled aerial videos, coupled with the challenges of collection and annotation, makes the task of training such modules more difficult. It turns out that the number and size of UAV video datasets is far fewer and smaller than those available for ground video datasets. For instance, the Kinetics dataset contains 650k videos, while the UAV-Human dataset has only 20k videos. Additionally, these learning-based methods are primarily designed for untrimmed videos, and adapting them to trimmed videos poses additional challenges.
% These methods typically train a learning-based module to select more informative frame sequentially or conditional early exiting in deterministic order. The performance of these methods heavily rely on the training data and can not be easily transfered to unseen action classes. However, it is harder to collect and annotate aerial videos and overall there are fewer and smaller UAV video datasets, as compared to ground video datasets (20k in UAV-Human vs 650k in Kinetics). Additionally, these methods are mainly designed for untrimmed videos, and adapting them in trimmed videos is hard due to the inherent difference between untrimmed and trimmed videos. 
Another alternative~\cite{Zhi2021MGSamplerAE} is to use a statistical model to represent the motion bias between frames, and devise an adaptive sampling strategy for frame selection based on motion information distribution along the temporal domain. However, this formulation fails to account for the distinct features of aerial video data, such as small resolution, multi-scale, and moving camera.
%Thus, the need arises for a more robust statistical formula that effectively leverages motion bias for improved frame selection in aerial videos.

\paragraph{Main Contribution:}
 We present a novel frame selection scheme for aerial action recognition. Our approach is general and is designed to address some of the challenges in aerial data. We utilize the similarity between patches to assess the amount of discriminative motion information contained in the aerial videos, and ensure that more informative frames can be selected for video representation. 
 Our method can be combined with any recognition model to obtain improved accuracy in terms of aerial action recognition. 
 The novel components of our work includes:

\begin{enumerate}
    \item We introduce patch mutual information (PMI) score to leverage the motion bias in the aerial videos. PMI score quantifies how much discriminative motion information is contained in one frame given another by measuring the similarity of frame patches via mutual information calculation.
    
    %We introduce patch mutual information (PMI) score to leverage the motion bias in the aerial videos. PMI score is used to represent the motion information between adjacent frames by measuring the similarity of frame patches via mutual information calculation. It quantifies how much discriminative motion information that is contained in one frame given another. PMI score provides an efficient yet robust way to distinguish the motion-salient frames from the aerial videos.
 
    %so that most of the pixels are well matched. Our approach is effective and robust to outliers.
    \item We propose an adaptive frame selection strategy based on shifted Leaky ReLu and cumulative distribution function. Our formulation enhances the motion bias between adjacent frames, making it easier to distinguish the motion-salient frames in videos. Furthermore, our method is designed in a direct plug-in manner to avoid complex training. 
    
    %We propose an adaptive frame selection strategy based on shifted leaky ReLu and Cumulative Distribution Function. It first maps the PMI score to $[0,1]$ and enhances the motion bias between adjacent frames, making it easier to distinguish the motion-salient frames in videos. It divides the whole video into multiple segments where each segment convey the same amount of motion information. Adaptive sampling is then performed by randomly picking a representative frame from each segment so that all the key information about the motion are included for training.
\end{enumerate}
% WHAT ARE DIFFERENT RECOGNITION MODELS THAT YOU HAVE INTEGRATED WITH?
We evaluate the effectiveness of our method in three aerial datasets and experimental results show that our approach consistently outperforms current state-of-the-art by large margins in terms of top-1 accuracy. Practically, we demonstrate a relative improvement of 2.2 - 13.8\% on UAV-Human~\cite{li2021uav}, 6.8\% on NEC Drone\cite{choi2020unsupervised}, and 9.0\% on Diving48 datasets\cite{li2018resound}.

\label{sec:intro}

%% file: figs/PMI_score.tex
% \begin{figure}[tp]
%     \centering
%     % \includegraphics[width=\linewidth]{}
%     \caption{Caption.}
%     \label{fig:template}
% \end{figure}

\begin{figure*}[tp]
    \centering

    \includegraphics[width=\textwidth]{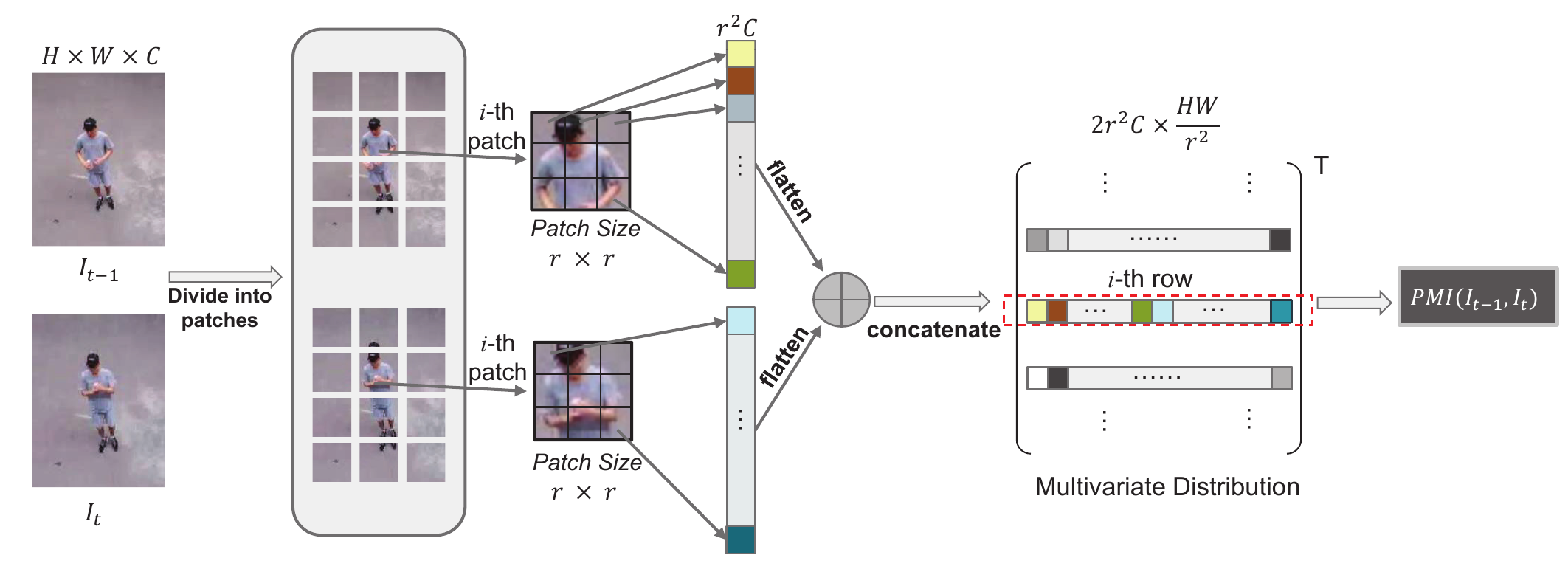}
    % \includesvg[width=\textwidth]{figs/PMI_score.svg}
    \vspace{-8mm}
    \caption{{\bf Patch Mutual Information formulation:} Given a pair of adjacent frames $I_{t-1}$ and $I_t$ in size $H\times W \times C$, we first divide them into patches with size $r \times r \times C$ and map each patch to a $d$-dimensional vector, where $d=r^2C$. We concatenate the two $d$-dimensional vectors representing the corresponding patches from two frames, the resulted vector is in $2d$ space. We perform the same operations for each patches and further concatenate these $2d$ vectors to get the multivariate distribution of size $2d \times \frac{HW}{r^2}$ for this image pair. Then, we calculate the covariance matrix of the multivariate distribution, compute the approximated entropies using Eq~\ref{eq:entropy_app} and eventually get the patch mutual information following the Eq~\ref{eq:overallmi}}.
    \label{fig:PMI} 
\vspace{-5mm}

\end{figure*}

%% file: 02_related.tex
\section{Related Work}

\subsection{Action Recognition for Aerial Videos}
The accuracy of action recognition on ground-camera video datasets has increased due to recent advancements in deep learning techniques. However, the current method cannot demonstrate a similar level of accuracy on videos captured using UAV cameras~\cite{nguyen2022state}. For aerial videos, \cite{geraldes2019uav},\cite{mliki2020human},\cite{mishra2020drone},\cite{mou2020event},\cite{barbed2020fine},\cite{gammulle2019predicting},\cite{mou2020event} utilize 2D convolutional neural networks (such as ResNet and MobileNet) as the foundational models for single-frame classification, and subsequently merge the outcomes of all video frames. Other methods~\cite{barekatain2017okutama},\cite{perera2019drone},\cite{perera2020multiviewpoint} utilize two-stream convolutional neural networks (CNNs) to leverage both human motion and appearance attributes for improved action recognition. \cite{choi2020unsupervised},\cite{demir2021tinyvirat},\cite{li2021uav},\cite{mou2020event},\cite{sultani2021human} employ the I3D network~\cite{carreira2017quo} to capture the spatial-temporal features of both human agents and their surroundings. \cite{aztr2023} proposes a framework leveraging CNNs and attention mechanisms for aerial action recognition on both edge devices and decent GPUs.  \cite{divya2022far, kothandaraman2023frequency} introduced an attention mechanism based on the Fourier transform to emphasize motion salience. Our proposed PMI Sampler is complimentary and could incorporate with aforementioned method to improve the overall recognition accuracy.

\subsection{Frame Sampling}
For some deep learning-based methods~\cite{carreira2017quo, Fan2018WatchingAS, Tran2014LearningSF}, the frame sequence used for training is obtained by randomly picking a fixed number of consecutive frames in the video. Other methods~\cite{feichtenhofer2020x3d,Wang2016TemporalSN} use a uniform sampling strategy, where frames are evenly sampled along the video's temporal domain. These two sampling techniques are commonly used for action recognition models. However, they do not exploit the motion bias between frames and do not consider the video characteristics corresponding to different human actions. 
%There are also some other sampling strategies were proposed recently.
There are also some learning-based frame selection methods~\cite{Wu2018AdaFrameAF,Korbar2019SCSamplerSS,gao2020listen,zheng2020dynamic,ijcai2018p98}. FastForward~\cite{ijcai2018p98} employs reinforcement learning for planning frame skipping and early stop decisions.
Adaframe~\cite{Wu2018AdaFrameAF} utilizes a policy gradient-trained LSTM, enhanced with a global memory for frame selection. SCSampler~\cite{Korbar2019SCSamplerSS} and Listen to Look~\cite{gao2020listen} utilize audio as an additional modality to exploit the natural semantic correlation between audios and frames. However, those methods mainly focus on untrimmed videos. They require large amounts of training samples and involve complex training procedures.

MGSampler~\cite{Zhi2021MGSamplerAE} leverages the temporal variations and use RGB difference between two adjacent frames to estimate the motion salience for each frame. However, such an approach may not be robust. Considering the challenges of aerial data (e.g. small resolution for human actors), their formulation may not be accurate due to the large number of outliers and noises belonging to the background information. In order to address these issues, we present a motion information representation technique using patch mutual information between frames and introduce a new sampling strategy based on the PMI score and shifted leaky ReLu. 

\subsection{Similarity Measure between Images}\label{subsec:similarity}
Numerous similarity metrics have been suggested for image analysis, with Euclidean distance being a common choice~\cite{zhi2021mgsampler}. Nevertheless, for UAV videos, Euclidean distance struggles due to pronounced background noise and frame instability. Cosine similarity is used for high-dimensional data but overlooks pixel value magnitude~\cite{hoe2021one}. Mutual information is a well-adopted similarity measure~\cite{Viola1995AlignmentBM,Maes1997MultimodalityIR}, especially in medical registration~\cite{Pluim2003MutualinformationbasedRO,Klein2007EvaluationOO}. It boasts resistance to outliers, yielding smooth cost functions for optimization~\cite{10.5555/1146355}. However, it lacks geometry consideration, solely focusing on pixel values, and neglecting spatial pixel relationships. NMI, which is a variation of Mutual Information, overlooks the spatial correlation between pixels and incurs higher computational costs. PSNR, derived from Mean Square Error (MSE), primarily focuses on pixel-level comparisons, is sensitive to dominant background changes encountered in aerial videos. Similarly, SSIM, a widely employed similarity measure, assesses the luminance, contrast, and structure of images but proves to be highly sensitive to structural variations such as rotations and shifts, which are frequently observed in aerial videos. Russakoff et al.~\cite{Russakoff2004ImageSU} propose RMI, integrating spatial data with mutual information, but it's computationally expensive. Inspired by these, we introduce patch mutual information, an RMI extension, simpler to implement yet provides enhanced accuracy.

\label{sec:related}

%% file: figs/Sampler.tex
\begin{figure*}[tp]
    \centering
    \includegraphics[width=\textwidth]{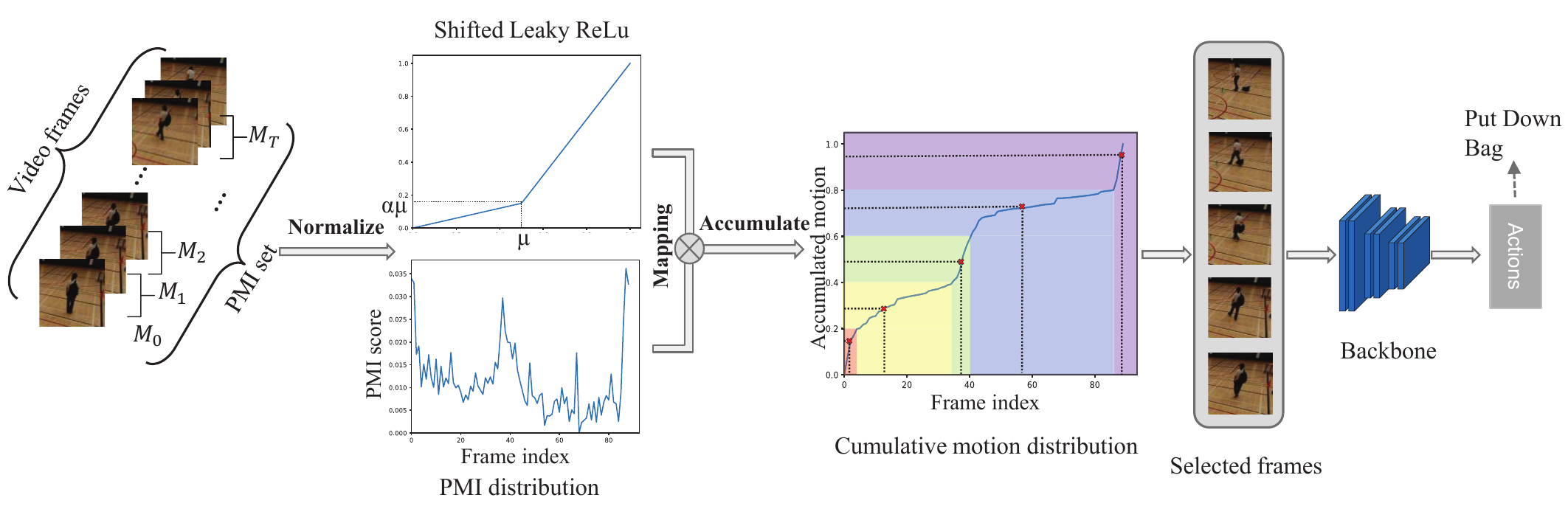}
    % \includesvg[width=\linewidth]{figs/ReLu.svg}
    \vspace{-7mm}
    \caption{ {\bf PMI Sampler:} Given a sequence of video frames, we compute the patch mutual information score for each pair of adjacent frames. We further remap the PMI scores using shifted Leaky ReLu to enhance the motion distribution. Finally, we accumulated the PMI scores using cumulative distribution function and segment the video into $N$ parts where $N$ is the number of samplings required for training. We randomly select one frame from each segments, constitute them into a sequence and feed them to the recognition backbones for action classification. }
    % and denote $M_t$ as the PMI between frames at time $t-1$ and $t$. For a video with $T$ frames, we then get a set of PMIs with length $T+1$, $M_0 = 0$. We first inversely map and normalize the PMIs to get the PMI scores $\hat{M_t}$, so that all the values are contrained betwen $[0,1]$ and higher value means there is more discriminative motion information contained in the current frame given the previous one. Moreover, we introduce shifted Leaky ReLu to remap the PMI scores and enhance the motion distribution, making it easier to distinguish the motion-salient frames. Finally, we accumulated the PMI scores using cumulative distribution function and segment the video into $N$ parts where $N$ is the number of samplings required for training. We randomly select one frame from each segments, constitute them into a sequence and feed them to the recognition backbones for action classification. }
    \label{fig:PMISAMPLER}
    \vspace{-4mm}
\end{figure*}

%% file: 03_method.tex
\section{Our Approach: PMI Sampler}
In this section, we present the details of our proposed patch similarity guided frame selection strategy. Specifically, we first introduce the concept of mutual information in Section~\ref{subsec:MI}. Next, we present the key component of our approach, including patch mutual information (PMI) score in Section~\ref{subsec:PMI_score}. Finally, we present the overall pipeline of our proposed PMI Sampler based on shifted leaky ReLu and cumulative distribution function in Section~\ref{subsec:ReLu}. 

\subsection{Mutual Information}\label{subsec:MI}
Mutual information (MI) is a concept in information theory that essentially measures the amount of information given by one variable when observing another variable. 
% It can also be interpreted as the reduction of the uncertainty of one variable given the other. 
Mutual information is highly correlated with entropy and joint entropy. The entropy is a measure of the uncertainty of a random variable and the joint entropy examines the overall complexity of all possible outcomes given both random variables. Specifically, given two discrete random variables $X$ and $Y$ with alphabet $\mathcal{X}$ and $\mathcal{Y}$. Their probability mass functions (PMFs) are denoted as $p_X(x)$ and $p_Y(y)$, the entropy of $X$,$H(X)$, and the joint entropy of $X$ and $Y$, $H(X,Y)$, can be calculated as:
\begin{equation}\label{eq:entropy}
 H(X) = -\sum_{x\in \mathcal{X}}p_X(x)\log p_X(x).   
\end{equation}
\begin{equation}\label{eq:jointentropy}
 H(X,Y) = -\sum_{x\in \mathcal{X},y\in\mathcal{Y}}p_{XY}(x,y)\log p_{XY}(x,y).  
\end{equation}
The mutual information between $X$ and $Y$ is defined as:
\begin{equation}\label{eq:overallmi}
\begin{split}
% I(X,Y) &= H(X)-H(X|Y) \\
%        &=H(Y)-H(Y|X) \\
     I(X;Y)  &=H(X)+H(Y)-H(X,Y), \\ 
\end{split}
\end{equation}
If we plug Eq~\ref{eq:entropy} and Eq~\ref{eq:jointentropy} into Eq~\ref{eq:overallmi}, we can get the calculation of mutual information as follows:
\begin{equation}\label{eq:miprob}
 I(X;Y)=\sum_{x\in\mathcal{X},y\in\mathcal{Y}}p_{XY}(x,y)\log\frac{p_{XY}(x,y)}{p_X(x)p_Y(y)}.
\end{equation}
The equation~\ref{eq:miprob} further suggests that mutual information essentially measures the distance between the real joint distribution $p_{XY}(x,y)$ and the distribution under the assumption of complete independence of $p_X(x)p_Y(y)$, which makes it a very nature measure of dependence~\cite{hyvarinen2000independent}. 

Mutual information can be used to measure the similarity between images~\cite{Viola1995AlignmentBM,Maes1997MultimodalityIR}. The probability distributions (in Eq~\ref{eq:miprob}) associated with images are normally approximated using respective marginal and joint histograms. One set of pixel co-occurrence in the joint distribution is represented by one entry in a two-dimensional joint histogram. 

Regional mutual information~\cite{Russakoff2004ImageSU} is an extension of mutual information. The intuition of RMI is to embed the spatial information of the images into mutual information calculations. Instead of standard pixel-wise mutual information, RMI represents a pixel as a multi-dimensional vector that consists of not only the pixel itself but also its neighboring pixels. However, the way that representing pixels with their neighborhood essentially makes extra focus on the local details. It may be beneficial for generic camera datasets, but considering that most of the pixels belong to the background in aerial data and the background details are redundant for action recognition, RMI is not a good choice to measure the motion information in aerial videos.

\subsection{Patch Mutual Information (PMI) score}\label{subsec:PMI_score}
% CAN YOU MATHEMATICALLY STATE THE PRECISE PROBLEM OF SELECTING THE DISCRIMINATIVE FRAMES HERE? IS THERE A SPECIFIC FUNCTION THAT YOU WANT TO OPTIMIZE IN TERMS OF FRAME SELECTION?

% YOU ARE SPENDING A LOT OF TEXT ON Zhi et al. AND RMI AS BACKGROUND IN THIS SECTION. THAT SHOULD BE IN SECTION 2 AND YOU SHOULD TALK ABOUT YOUR APPROACH HERE.

% Our goal is to select the more discriminative frames which contain more motion information about the action. Therefore, we need an relatively accurate representations for motion information along the temporal domain. CAN YOU MATHEMATICALLY STATE WHAT DOES THIS MOTION INFORMAITON CORRESPONDS TO?
% Zhi et al.~\cite{zhi2021mgsampler} SHALL YOU TALK ABOUT ZHI ET AL IN SECTION 2 OR LATER FOR COMPARISON? WHY IN THIS SECTION? leverage the temporal variations and use RGB difference between two adjacent frames to estimate the motion salience for each frame. However, such representation is not robust and considering the challenges of aerial data (e.g. small resolution for human actor), their formulation may not be accurate due to the large number of outliers and noises belonging to the background information. 
% %It lacks the capability to represent the motion information between frames. 
% To solve such issue, we use concepts from information theory and treat the images as signals. We introduce patch mutual information score (PMI) to quantify the motion bias between adjacent frames by measuring the similarity of patches. PMI assesses the amount of discriminative motion information contained in one frame, given the other frame.

Inspired by regional mutual information (RMI), we propose patch mutual information (PMI). Instead of mapping pixels into multi-dimensional space, we divide the image into patches and map each patch into a multi-dimensional points and then calculate the mutual information between corresponding patches in adjacent frames. 
% PMI can also be interpreted as a sparse version of RMI. Contrary to representing each of the pixels into $d$-dimensional points, where $d = r^2$ and $r$ is square length of the patch, we now sparsely map the pixels with a fixed stride of $r$. 
In this manner, we can encode the spatial relationships of patches in the mutual information calculation without giving too much focus on the redundant background noises, which yields to better motion information representation for aerial videos.

\noindent \textbf{Entropy approximation} We can exactly calculate the patch mutual information in the traditional manner that mentioned in Section~\ref{subsec:MI}, but now with $d$-dimensional histograms for marginal probability distribution and $2d$-dimensional histograms for joint distribution. However, such operation seems reasonable but is hard to implement in practice. In fact, we can make the calculation easier by assuming that the multivariate distribution of the image is normally distributed~\cite{Russakoff2004ImageSU} and it is well supported by the $m$-dependence variable concept proposed in ~\cite{10.1215/S0012-7094-48-01568-3}.

For a normally distributed set of points $P=[p_1,p_2,\cdots, p_N]^T$, $p_1,p_2,\cdots, p_N \in R^d$, the entropy can be calculated as:
\begin{equation}\label{eq:entropy_app}
 H(P) = \frac{1}{2}log((2\pi e)^d det(var(P))).   
\end{equation}
where $var(P)$ is the covariance matrix of $P$. Let $E(P)$ be the expected value of $P$, i.e, the mean of $P$, covariance matrix of $P$ is defined as:
\begin{equation}\label{eq:var}
var(P) = E[(P-E[P])(P-E[P])^T],  
\end{equation}

% In practice, given a multivariate distribution represented by a set of points $P=[p_1,p_2,\cdots,p_d]^T$, we can use the same process that is used for principal components analysis(PCA)~\cite{inbook} to calculate this approximation TO WHAT QUANTITIY.
% by first subtracting the mean from the points and diagonalizing its covariance matrix and eventually summing the entropies along each dimension.

\noindent \textbf{Patch mutual information formulation} We here describe the detailed procedures to get the multivariate distribution for a pair of images and calculate the patch mutual information between these two images.

As shown in Figure~\ref{fig:PMI}, given frame $I_{t-1}$ and $I_{t}$, we divide each frame into patches with square length $r$. For each patch pair, we flatten the patches into vectors in $d$-dimensional space where $d=Cr^2$ and concatenate them to a vector in $R^{2d}$. For each pair of frames in $R^{H \times W \times C}$, we will have $N=\frac{HW}{r^2}$ patches (we simply ignore the pixels along the edges as they will not have a significant effect on the final entropy). Since each pair of patches has been mapped into a $2d$ vector, we then have a multivariate distribution of those $N$ points by further concatenating the $2d$ vectors into a $2d \times N$ matrix, $P_t=[p_1,p_2,\cdots ,p_N]$.

We calculate the covariance matrix of $P_t$ by first centering all the $N$ points to their means and then compute the covariance of $P_t$ following Eq.~\ref{eq:var}. The covariance matrix of $P_t$ is denoted as $\Omega$:
\begin{equation}\label{eq:cov}
\Omega = \frac{1}{N}(P_t - \frac{1}{N}\sum^N p_i)( P_t - \frac{1}{N}\sum^N p_i)^T.
\end{equation}

We estimate the entropy of these points $H(P_t)$ using Eq~\ref{eq:entropy_app}. Since we concatenate the patches from two frames to form $P_t \in R^{2d \times N}$, $H(P_t)$ is essentially the approximation of the joint entropy between frame $I_{t-1}$ and $I_t$. However, we can easily obtain the covariance matrices of each frame. Given that $P_t$ is a combination of two $d$-dimensional vectors, the computation of the covariance matrix of $P_t$ also results in the covariance matrices of these 2 subsets of points. In practice, the covariance of frame $I_{t-1}$, denoted as $\Omega_{t-1}$, corresponds to the $d \times d$ matrix in the top left of $\Omega$ and $\Omega_t$ is the $d \times d$ matrix in the bottom right representing the covariance of frame $I_t$. Moreover, the marginal entropies, $H(\Omega_{t-1})$ and $H(\Omega_t)$, can be computed using the same formulation in Eq~\ref{eq:entropy_app}.

Finally, we obtain the patch mutual information in a pair of frames $I_{t-1}$ and $I_t$, denoting as $M_t$, following the Eq~\ref{eq:overallmi}:
\begin{equation}\label{eq:PMI_final}
M_t = PMI(I_{t-1},I_t) = H(\Omega_{t-1}) + H(\Omega_t) - H(\Omega).
\end{equation}

% Given an image of size $H \times W \times C$, the calculation complexity of RMI has complexity $O(4HWC^2r^4))$, while our proposed PMI's complexity is $O(4HWC^2r^2)$. 

\noindent \textbf{Patch mutual information score} 
% For a video with $T$ frames, we sequentially compute the PMI between frame $I_{t-1}$ and frame $I_{t}$ and denote that as $M_t$. 
Note that, PMI holds the properties of standard mutual information and is always greater than or equal to 0. In case, the two frames are complete independent from each other, then $PMI=0$. However, since video contents are always consistent and have some coherence between the frames, the two successive frames are always correlated and dependent. Therefore, PMI between adjacent frames will always be greater than zero, $M_t>0$ for $t \in {1,2,\cdots,T}$. For frame at time 0, we simply define $M_0=0$. 

When the two frames are more discriminative, the dependence relationship weakens and PMI gets smaller, indicating there is potentially more motion information contained in the current frame. The scale and range of the PMI in different videos also varies. To better analyze the motion information distribution, we inversely remap the PMI $M_t$ and normalize it with the $l_1$ norm:
\begin{equation}\label{eq:inversemapping}
M^{\prime}_t = \max_{i\in T}(M_i) - M_t,
\end{equation}
\begin{equation}\label{eq:inversenormalize}
\hat{M}_t = \frac{M^{\prime}_t}{\sum_T M^{\prime}_t}.
\end{equation}
Eventually, we obtain the PMI score $\hat{M}_t$. PMI score represents the discriminative motion information contained in current frame given the previous frame. The higher PMI score indicates there existing salient motion in current frame.

\subsection{PMI Sampler: Frame Selection}\label{subsec:ReLu}
We present a novel frame selection strategy based on shifted leaky ReLu and cumulative distribution function (CDF). The intuition behind our method is to sample a sequence of frames that contains as much discriminative motion information as possible. Similar to MG Sampler~\cite{zhi2021mgsampler}, our method also utilize a temporal segmentation scheme and adaptively selects frames according to the motion information distribution over the entire video.

% Since the motion in trimmed videos is normally consistent, the deviations in the motion information distribution tends to be marginal.  YOU SHOULD TALK ABOUT PRIOR WORK IN SECTION 2 AND NOT HERE. MG Sampler~\cite{Zhi2021MGSamplerAE} uses cumulative distribution function and uses a hyper-parameter to adjust the original motion distribution, but their formulation does not work very well in aerial videos. 
%%%%%%%%%%%%%%%%%%%%%%%%%%%%%%%%%%%%%%%%%%%%%%%%
% Because aerial videos are normally captured using changing camera views, the MG's L2 distance is too sensitive. And their sqrt remap operation will tend to inhibit the larger values, which refer to the important frames. THIS COMPARISON WIHT MG SAMPLER SHOULD BE IN SECTION 2 (PRIOR WORK) OR SECTION 4 (COMPARISONS).
%%%%%%%%%%%%%%%%%%%%%%%%%%%%%%%%%%%%%%%%%%%%%%%%%%
 % Instead, we propose to use a shifted leaky ReLu to remap the PMI score and enhance the motion bias so that it is easier to distinguish the frames that contain more useful motion information.

As shown in Figure~\ref{fig:PMISAMPLER}, given a video contain $T$ frames, we calculate PMI between adjacent frames and generate a set of PMI scores, $S = \{\hat{M_0},\hat{M_1},\cdots,\hat{M_T}\}$ with mean $\mu$. For each element $\hat{M_t}\in [0,1], t\in \{0,1,\cdots,T\}$, we map it based on the following equation:
\begin{equation}\label{eq:leaky}
M^*_t = \left\{
\begin{array}{lcl}
\alpha \hat{M_t} & & {\hat{M_t} \leq \mu},\\
\frac{1-\alpha\mu}{1-\mu}(\hat{M_t}-1)+1 & & {\hat{M_t} > \mu}.
\end{array} \right .
\end{equation}
where $\alpha$ is a hyper parameter to control the smoothness. This function is very much similar to Leaky ReLu, but with the origin shifts to $(\mu,\alpha\mu)$ and the function is constrained by the origin and $(1,1)$. Therefore, we call it as {\em shifted Leaky ReLu}. As shown in Figure~\ref{fig:Relu}, it polarizes the PMI scores and makes the motion-salient frames more distinct. We further examine the impact of $\alpha$ in Section~\ref{subsec:ablation}.

Then, we normalize $M^*_t$ again using $l_1$ norm and accumulate the the remapped PMI score to get the cumulative motion distribution of the video:
\begin{equation}\label{eq:cdf}
 F_T(t) = \sum_{i\leq t}M^*_i.
\end{equation}
The constructed cumulative distribution of motion information is shown in Fig~\ref{fig:PMISAMPLER}, where the $X$-axis represents the frame index and $Y$-axis stands for the motion information accumulation up to the current frame. 

Finally, we divide the video frames into N parts with interval $1/N$ on the $Y$-axis. The frames with index in the corresponding interval on the $X$-axis will be clustered as one segment. Considering that the $X$-axis value may not be an integer, we choose the closest integer value instead. Then, we will randomly select one frame from each segment and constitute a frame sequence to represent the video for the recognition. 
As shown in Fig~\ref{fig:PMISAMPLER} and Fig~\ref{fig:example}, our approach can select more frames during periods with higher motion salience and less frames in motion static segments, enabling more discriminative motion information to be conveyed to the recognition model.

% THIS MOTIATION SHOULD BE PRESENTED FIRST, BEFORE LISTING THE ACTUAL FORMULA BASED ON RELU?
% The motivation of the remapping using shifted leaky ReLu is to polarize the motion bias between frames so that the motion-salient frames become more distinct. 
% % need more details

% ARE YOU GOING TO DESCRIBE THESE POINTS IN PROPER BULLETS AND SENTENCES.

% 1. PMI robust -- nonsensitive to noise
% 2. polarization -- detial action change -- more valid frames
% 3. large motion with more information -- steep -- less frames to choose  -- train more -- right focus
% %
%This is reasonable since one action can always be divided into several stages and the relatively high motion bias means the action gets into the next stage. The number of frames belonging to motion salient segments is smaller, however, those frames are usually in the stage transition period. Our method could enable the recognition model repeatedly revisit those frames and learn from the discriminative motion information about the action. 
%
 % CAN YOU SHOW THIS COMPARISON USING SOME VISUALIZATION IN THE SUPPLEMENTARY MATERIA

% However, owing to the extremely rare viewing angles (oblique and overhead), the fluctuation of the motion distribution is much harder to notice in aerial videos.
% Next step, we want to utilize the PMI score to help us better distinguish the motion-salient frames in the videos, so that we can have better understanding on the motion information distributions. 

\label{sec:method}

%% file: figs/ReLu.tex
\begin{figure}[t]
    \centering
    \includegraphics[width=\columnwidth]{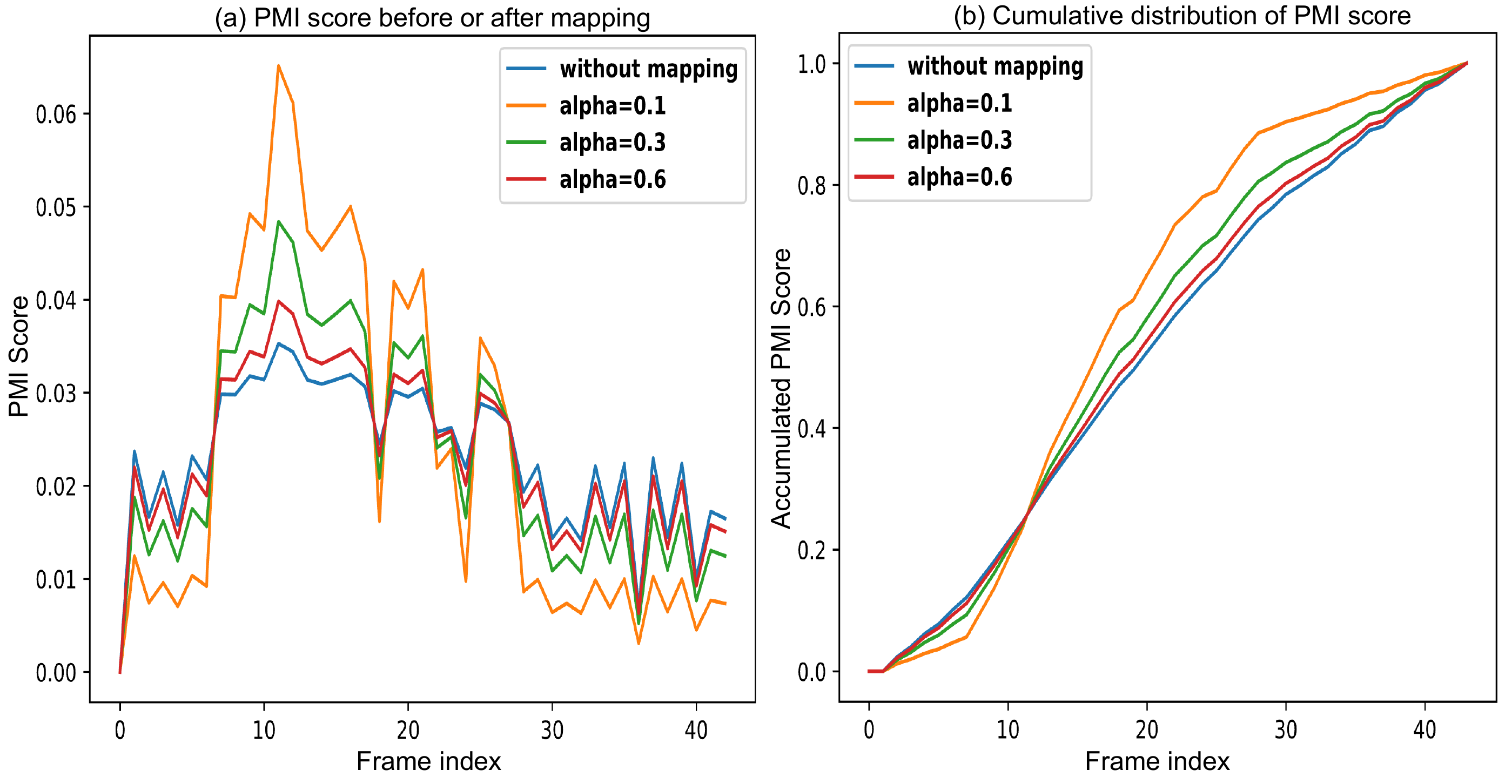}
    % \includesvg[width=\linewidth]{figs/ReLu.svg}
    \caption{The PMI score and cumulative distribution before or after the shifted Leaky ReLu mapping under different value of $\alpha$. After mapping, motion salient frames become more distinct.}
    \label{fig:Relu}
\vspace{-5mm}
\end{figure}

%% file: 04_results.tex
\section{Results}
\label{sec:results}

In this section, we present the experimental results of our approach along with the state-of-the-art methods on three aerial datasets: UAV-Human~\cite{li2021uav}, NEC Drone~\cite{choi2020unsupervised} and Diving48~\cite{li2018resound}. We also conduct ablation studies on the effects of patch mutual information, shifted leaky ReLu, and patch size. Our approach can incorporate with any existing recognition backbones, we use X3D~\cite{feichtenhofer2020x3d} as our default recognition backbone model unless otherwise specified. The implementation details are included in supplementary material.

\subsection{Results on UAV-Human}
UAV Human~\cite{li2021uav} is regarded as the largest and most comprehensive dataset of UAV-based human behavior understanding data to date. The collection includes 22,476 high-definition videos captured in various indoor and outdoor settings, encompassing a broad range of lighting and weather conditions. The videos showcase dynamic backgrounds and feature diverse UAV motions and flying altitudes, making this dataset highly challenging. A total of 155 unique actions have been annotated, with some being difficult to differentiate, such as squeeze and yawn.

We evaluate the performance of our proposed PMI Sampler on UAV-Human along with current state-of-the-art methods. The results are shown in Table~\ref{tab:uavhuman}. All the videos are pre-processed with the same procedures in MITFAS~\cite{mitfas2023}, and followed by data augmentation that consistent with X3D~\cite{feichtenhofer2020x3d}. Our method demonstrates a relative improvement in top-1 accuracy over the current state-of-the-art methods by $2.2 - 13.8\%$, with regards to different settings in number of frames, input frame size, and model initialization.

\input{tables/uavhuman}
\input{tables/diving48}
\subsection{Results on Diving48 and NEC-Drone}
NEC Drone~\cite{choi2020unsupervised} is an indoor video dataset that features 5,250 videos depicting 16 distinct actions performed by 19 actors. The videos were captured using a UAV flying at low altitude over a basketball court. In contrast to the UAV Human dataset, the lighting conditions in NEC Drone are more consistent; however, the dataset is plagued by noise due to light reflections. Diving48~\cite{li2018resound} is a comprehensive video dataset that offers a fine-grained analysis of competitive diving, free from any significant biases towards static or short-term motion representations. It comprises approximately 18,000 trimmed video clips, each depicting one of 48 unambiguous dive sequences. Although it is not a UAV-captured dataset, the majority of the videos are captured by cameras at high altitudes and oblique angles with dynamic movement.

We compare our method with other state-of-the-art methods on Diving48 and NEC-Drone. The frames are extracted from raw videos and augmented as in X3D~\cite{feichtenhofer2020x3d}. The baseline methods are uniform and random samplings. As shown in Table~\ref{tab:diving}, on diving48, our method achieves 10.6 - 14.3\% relative improvement in top-1 accuracy over the baseline methods and relative 9.0\% over the SOTA. On NEC Drone, PMI Sampler outperforms the baseline methods by 12.8 - 20.2\% and improves by 6.8\% over the SOTA, relatively.

\input{tables/similarity_measure}
\input{tables/mapping_TABLE}
\subsection{Ablation Studies}\label{subsec:ablation}
We conduct ablation studies on the effectiveness of patch mutual information and shifted leaky ReLu. We also explore the impact of the hyper-parameter $\alpha$ in the shifted leaky ReLu functions as well as the size of the patches. We generate subsets of UAV-Human and Diving48 by randomly choosing 30\% videos per class from the original datasets and denote them as UAV-Human subset ($\sim$ 6k videos) and Diving48 subset ($\sim$ 5.8k videos).  

\noindent \textbf{Effectiveness of patch mutual information:} As mentioned in Section~\ref{subsec:similarity}, there are many other similarity measurements like Euclidean distance, cosine similarity, . We conduct experiments to test the results generated based on different similarity measures.  As shown in table~\ref{tab:similarity}, our proposed PMI outperforms other similarity measures.

\noindent \textbf{Effectiveness of shifted Leaky ReLu:} After getting the PMI score for the video, we remap it using shifted Leaky ReLu. It polarizes the motion information distribution and makes it easier to distinguish the motion-salient frames in aerial videos, see Figure~\ref{fig:Relu}. However, there are many other mapping functions we can use, like sigmoid, softmax, tanh functions. We conduct ablation experiments on different mapping functions, and the results are shown in Table~\ref{tab:mapping}, which demonstrate that shifted Leaky ReLu is the better choice for aerial videos.

\noindent \textbf{Impact of the hyper-parameter in shifted leaky ReLu:} We also conduct experiments to explore the impact of the hyper-parameter $\alpha$ in the shifted leaky ReLu function. Results are shown in Table~\ref{tab:hyper}. When $\alpha$ is smaller, the distance between small PMI scores and large PMI scores will be enlarged, yielding to steeper cumulative motion information distribution, see Figure~\ref{fig:Relu}. With $\alpha$ larger, such distance will be reduced and the cumulative distribution will be smoother.

\input{tables/lessinfo}
\input{tables/hyper}

\noindent \textbf{Impact of the patch size:} We compare the results from PMI scores that were generated using different sizes of patches. As we mentioned in Section~\ref{subsec:PMI_score}, the entropy approximation is based on the assumption that the multivariate distribution of the image is normally distributed, and it is well supported by the $m$-dependence variable concept. Such approximation is closer to the real entropy when the patch size is larger. However, if the patch size is larger, the time for the calculation will be more expensive. As shown in Table~\ref{tab:dimension}, we find out that when patch size is $7\times 7$, we achieve the best trade-off between accuracy and time cost. Therefore, we divide the images into  $7\times 7$ patches for all the other experiments in this paper unless further specified.

\input{tables/dimension}
\input{tables/sthsth}

\noindent \textbf{Training efficiency:} The training time on the Diving48 dataset is presented in Table~\ref{tab:lessinfo}. Similar to the MG Sampler, our approach computes the PMI in the pre-processing phase to avoid redundant computations for the same set of frames during training. 
PMI Sampler achieves improved accuracy without significantly increasing the training time. Also, compared to MG Sampler, our proposed PMI Sampler achieves better accuracy with less spatial or temporal information (fewer frames and smaller input size).

\noindent \textbf{Results on ground datasets:} We evaluate our proposed PMI Sampler on general ground camera datasets like SomethingSomething V2, UCF101 and HMDB51. UCF101 and HMDB51 both have 3 train-test splits, we report the average results here with X3D as the backbone. As shown in Table~\ref{tab:sth}, PMI Sampler also outperforms the current state-of-the-art on these datasets.

More ablation studies are shown in Appendix.~\ref{sec:back} and \ref{sec:dense}. We also include a detailed analysis in Appendix.~\ref{sec:analysis} and more visualization results in Appendix.~\ref{sec:visualresults}.

% Therefore, PMI is robust to background noise and takes both pixel value and pixel position into account, making it a perfect motion information representation for aerial videos.

%% file: tables/uavhuman.tex
\begin{table}[!t]
\centering
\resizebox{1.0\columnwidth}{!}{
\begin{tabular}{c c c  c c}
\toprule
Method & \makecell{Frames \\ Number} & Input Size & Init. & \makecell{Top-1 (\%)  \\ Acc. $\uparrow$}   \\
\midrule 
% X3D-S \cite{feichtenhofer2020x3d}& - & $16$ & $224\times224$ & None & $21.5$ \\
% X3D-M \cite{feichtenhofer2020x3d}& - & $16$ & $224\times224$ & None & $27.0$ \\
% X3D-L \cite{feichtenhofer2020x3d}& - & $16$ & $224\times224$ & None & $27.6$ \\
% FAR \cite{divya2022far} & X3D-M & $16$ & $224\times 224$ & None & $27.6$ \\
% MITFAS & X3D-M & $16$ & $224\times 224$ & None & 40.2 \\
% \textbf{Ours} & X3D-M & $16$ & $224\times 224$ & None & \textbf{X} \\
% \midrule 
FAR \cite{divya2022far} & $8$ & $540\times 540$ & None & $28.8$ \\
MITFAS \cite{mitfas2023} & $8$ & $540\times 540$ & None & 38.4 \\
\textbf{Ours} & $8$ & $540\times 540$ & None & \textbf{39.7} \\
\midrule 
% I3D \cite{carreira2017quo}& ResNet-101 & $8$ & $540\times960$ & Kinetics & $21.1$ \\
% FNet \cite{lee2021fnet} & I3D & $8$ & $540\times960$ & Kinetics & $24.3$ \\
% FAR \cite{divya2022far} & I3D & $8$ & $540\times960$ & Kinetics & $29.2$ \\
X3D-M  \cite{feichtenhofer2020x3d}& $8$ & $540\times540$ & Kinetics & $36.6$ \\
FAR \cite{divya2022far} & $8$ & $540\times540$ & Kinetics & $38.6$ \\
DiffFAR \cite{kothandaraman2022differentiable} & $8$ & $540\times540$ & Kinetics & $41.9$ \\
\textbf{Ours} & $8$ & $540\times 540$ & Kinetics & \textbf{47.7} \\
\midrule 
FAR \cite{divya2022far}& $8$ & $620\times620$ & Kinetics & $39.1$ \\
MITFAS \cite{mitfas2023} & $8$ & $620\times 620$ & Kinetics & $46.6$ \\
\textbf{Ours} & $8$ & $620\times 620$ & Kinetics & \textbf{52.0} \\
\midrule 
X3D-M  \cite{feichtenhofer2020x3d}& $16$ & $224\times224$ & Kinetics & $30.6$ \\
FAR \cite{divya2022far} & $16$ & $224 \times 224$ & Kinetics & $31.9$ \\
AZTR \cite{aztr2023} & $16$ & $224 \times 224$ & Kinetics & $47.4$ \\
MITFAS \cite{mitfas2023}  & $16$ & $224\times 224$ & Kinetics & 50.8 \\
MG Sampler \cite{zhi2021mgsampler}  & $16$ & $224\times 224$ & Kinetics & 53.8 \\
\textbf{Ours} & $16$ & $224\times 224$ & Kinetics & \textbf{55.0} \\
\bottomrule 
\end{tabular}
}
\vspace{-7pt}
\caption{\textbf{Results on UAV-Human.} We demonstrate relative improvements in the top-1 accuracy by $2.2 - 13.8\%$ over previous state-of-the-art methods. Our method outperforms the state-of-the-art methods under different settings, which further indicates the benefits of our proposed PMI Sampler.}
\vspace{-5pt}
\label{tab:uavhuman}
\end{table}

%% file: tables/diving48.tex
\begin{table}
\small
\centering
% \begin{center}
\resizebox{1.0\columnwidth}{!}{
\begin{tabular}{c c c c  c}
\toprule
Method & Frames & Backbone  & \makecell{Diving48\\ Top-1 (\%)} & \makecell{NEC Drone\\ Top-1 (\%)}   \\
\midrule
Random \cite{feichtenhofer2020x3d}&  $16$& X3D-M  & $71.1$ & $52.0$ \\
Uniform \cite{feichtenhofer2020x3d}&  $16$& X3D-M  & $73.5$ & $55.4$\\
MG Sampler \cite{zhi2021mgsampler} &  $16$& X3D-M  & $74.6 $ & $58.5$\\
K-centered \cite{park2022k} &  $16$& ViT~\cite{dosovitskiy2020image} & $72.5$ & $36.3$\\
\midrule
\textbf{Ours} &  $16$& X3D-M  & \textbf{81.3} & \textbf{62.5} \\

% FAR \cite{divya2022far} &  $16$ & $224\times224$ & Kinetics & $71.4$ \\
% %Diff-FAR \cite{kothandaraman2022fourier} &  $8$ & $960\times540$ & Kinetics & $80.75$ \\
% MITFAS &  $16$ & $224\times224$ & Kinetics & $78.6$ \\
% \textbf{Ours} &  $16$ & $224\times224$ & Kinetics & \textbf{78.6} \\
\bottomrule
\end{tabular}
}
\vspace{-7pt}
\caption{\textbf{Results on Diving48 and NEC-Drone.} Our method relatively improves the top-1 accuracy by 9.0\% on Diving48, by 6.8\% on NEC Drone. }
\vspace{-10pt}
\label{tab:diving}
% \end{center}
\end{table}

%% file: tables/similarity_measure.tex
\begin{table}
% \footnotesize
\centering
% \begin{center}
\resizebox{1.0\columnwidth}{!}{
\begin{tabular}{c c c}
\toprule
Similarity measure & \makecell[c]{UAV-Human subset \\ Top-1 Acc (\%)} & \makecell[c]{Time cost \\ per frame (ms)}   \\
\midrule
% l1 norm & $54.1$ & -  \\
Euclidean Distance &  $56.4$ &  $1.7$\\
Cosine Similarity & $55.0$ & $2.5$\\
Mutual Information &$58.4$ & $10.6$\\
Regional Mutual Information (RMI) & $59.2$ & $20.2$\\
Normalized Mutual Information (NMI) &  58.8 &  32.8 \\
Peak Signal-to-Noise Ratio (PSNR) & 57.0 & 3.7 \\
Structural Similarity Index Measure (SSIM) & 57.7 & 79.9 \\
\textbf{Patch Mutual Information} &  \textbf{59.8} & \textbf{4.5} \\

% FAR \cite{divya2022far} &  $16$ & $224\times224$ & Kinetics & $71.4$ \\
% %Diff-FAR \cite{kothandaraman2022fourier} &  $8$ & $960\times540$ & Kinetics & $80.75$ \\
% MITFAS &  $16$ & $224\times224$ & Kinetics & $78.6$ \\
% \textbf{Ours} &  $16$ & $224\times224$ & Kinetics & \textbf{78.6} \\
\bottomrule
\end{tabular}
}
\vspace{-7pt}
\caption{\textbf{Comparison between different similarity measures}. Our proposed Patch Mutual Information(PMI) outperforms other measures on UAV-Human.}
\vspace{-5pt}
\label{tab:similarity}
% \end{center}
\end{table}

%% file: tables/mapping_TABLE.tex
\begin{table}
% \footnotesize
\centering
% \begin{center}
\resizebox{1.0\columnwidth}{!}{
\begin{tabular}{c c c}
\toprule
Mapping function & \makecell{NEC Drone \\ Top-1 Acc (\%)} & \makecell{Diving48 Subset\\ Top-1 Acc (\%)}  \\
\midrule
without Mapping & $59.3$ & $58.8$ \\
Quadratic & $60.3$ & $63.4$ \\
Sigmoid &  $61.4$ &  $65.7$ \\
Softmax & $58.4$ & $53.9$  \\
Tanh & $60.9$ & $56.6$\\
ReLu & $60.5$ & $60.6$\\
\textbf{Shifted Leaky ReLu} & \textbf{62.5} & \textbf{66.3}\\

% FAR \cite{divya2022far} &  $16$ & $224\times224$ & Kinetics & $71.4$ \\
% %Diff-FAR \cite{kothandaraman2022fourier} &  $8$ & $960\times540$ & Kinetics & $80.75$ \\
% MITFAS &  $16$ & $224\times224$ & Kinetics & $78.6$ \\
% \textbf{Ours} &  $16$ & $224\times224$ & Kinetics & \textbf{78.6} \\
\bottomrule
\end{tabular}
}
\vspace{-7pt}
\caption{\textbf{Comparison between different mapping functions.} Shifted Leaky ReLu can better map the motion information distribution and make it easier to distinguish frames containing more motion information.}
\vspace{-10pt}
\label{tab:mapping}
% \end{center}
\end{table}

%% file: tables/lessinfo.tex
\begin{table}
\small
\centering
% \begin{center}
\resizebox{1.0\columnwidth}{!}{
\begin{tabular}{c c c c c  c c}

\toprule
Method & Frames & Input Resolution  & \makecell[c]{Training Time \\ per Epoch (s) $\downarrow$}& \makecell[c]{Training Time \\ per Video (ms) $\downarrow$}& \makecell[c]{ Diving48 \\ Top-1(\%)} $\uparrow$&  \makecell[c]{Diving48 \\ Top-5(\%)} $\uparrow$   \\
\midrule
Random &  $16$ & $224 \times 224$ &  196.6 & 13.1 & $71.1$ & $94.8$\\
Uniform &  $16$ & $224 \times 224$ &  191.4 &12.7& $73.5$ & $95.1$\\
MG Sampler  &  $16$ & $224 \times 224$ & 287.4 &19.2& $74.6 $ & $95.0$\\
\midrule
PMI Sampler $\textbf{(Ours)}$ &  $\textbf{8}$ & $224 \times 224$ &  248.3 $\textbf{(-39.1)}$ & 16.5 $\textbf{(-2.7)}$ & 75.0 $\textbf{(+0.4)}$ & 95.6 $\textbf{(+0.6)}$ \\
PMI Sampler $\textbf{(Ours)}$ &  $\textbf{12}$ & $224 \times 224$ &  276.8 $\textbf{(-10.6)}$ & 18.4 $\textbf{(-0.8)}$ & 77.7 $\textbf{(+3.1)}$ & 97.5 $\textbf{(+2.5)}$ \\
PMI Sampler $\textbf{(Ours)}$ &  $16$ & $\textbf{172 $\times$ 172}$ &  294.8 $\textbf{(+7.4)}$ & 19.6 $\textbf{(+0.4)}$ &79.0 $\textbf{(+4.4)}$ & 97.7 $\textbf{(+2.7)}$ \\
\midrule
PMI Sampler $\textbf{(Ours)}$ &  $16$ & $224 \times 224$ &  295.1 $\textbf{(+7.7)}$ & 19.6 $\textbf{(+0.4) }$ & 81.3 $\textbf{(+6.7)}$ & 97.7 $\textbf{(+2.7)}$ \\
\bottomrule
\end{tabular}
}
\vspace{-7pt}
\caption{\textbf{Training efficiency.} PMI Sampler achieves higher accuracy with less spatial/temporal information, leading to better speed-accuracy tradeoffs. We show results for fewer frames (8 and 12) and smaller input sizes (172$\times$172).}
\vspace{-5pt}
\label{tab:lessinfo}
% \end{center}
\end{table}

%% file: tables/hyper.tex
\begin{table}
% \footnotesize
\centering
% \begin{center}
\resizebox{1.0\columnwidth}{!}{

\begin{tabular}{cccccccccc}
\toprule
\multicolumn{1}{c}{} & $\alpha$ & 0 & 0.1 & 0.2 & \textbf{0.3} & 0.4 & 0.5 & 0.6 & 0.7 \\ \midrule
\multirow{2}{*}{\makecell{Diving48 \\subset}}   
& Top-1 (\%) & 60.6 & 65.4 & 64.7 & \textbf{66.3} & 66.2 & 65.4 & 65.6 & 65.4 \\
& Top-5 (\%) & 92.6 & 92.8 & 94.4 & \textbf{94.9} & 94.6 & 93.3 & 94.5 & 94.6 \\ 
\midrule
\multirow{2}{*}{\makecell{UAV-Human \\subset}}    
& Top-1 (\%) $\uparrow$ & 53.7 & 58.1 & 59.8 & \textbf{59.8} & 59.7 & 58.7 & 59.5  & 59.3\\
& Top-5 (\%) $\uparrow$ & 83.8 & 85.3 & \textbf{86.5} & 86.3 & 85.3 & 86.2 & 85.5 & 86.0\\ 
\midrule
\multirow{2}{*}{\makecell{NEC-Drone}}    
& Top-1 (\%) $\uparrow$ & 54.1 & 57.9 &58.0 & 58.2 & 60.7 & 61.2 & \textbf{62.5}  & 61.7\\
& Top-5 (\%) $\uparrow$ & 85.4 & 89.5 & 88.6 & 89.3 & 89.5 & \textbf{90.0} & 89.6 & 89.5\\ 
\bottomrule

\end{tabular}

% \begin{tabular}{l l l l l l l l l}
% \toprule
% \multicolumn{1}{c}{} & $\alpha$ & 0 & 0.1 & 0.2 & 0.3 & 0.4 & 0.5 & 0.6    \\
% \midrule
% \multirow{2}{c}{Diving48 subset} & Top-1 & 60.6 & 65.4 & 64.7 & 66.3 & 66.2 & 65.4 & 65.6 \\
%                                  &Top-5 & 92.6 & 92.8 & 94.4 & 94.9 & 94.6 & 93.3 & 94.5\\

% \bottomrule
% \end{tabular}
}
\vspace{-7pt}
\caption{\textbf{Impact of the hyper-parameter $\alpha$.} The dynamic camera movement in UAV-Human and Diving48 videos necessitates a lower $\alpha(\sim0.3)$ to differentiate real motion from background noise. In contrast, NEC-Drone videos, captured by a hovering UAV, require a higher $\alpha(\sim0.6)$ to preserve the motion info distribution.  }
\vspace{-10pt}
\label{tab:hyper}
% \end{center}
\end{table}

% \begin{table}
% % \footnotesize
% \centering
% % \begin{center}
% \resizebox{0.7\columnwidth}{!}{
% \begin{tabular}{c c c c c c c}
% \toprule
% $\alpha$ & Top-1 Accuracy(\%) & Top-5 Accuracy(\%)   \\
% \midrule
% 0 & $60.63$ & $92.62$ \\
% 0.1 & $65.38$ & $92.79$  \\
% 0.2 & $64.67$ & $94.38$ \\
% 0.3 &  $66.26$ &  $94.9$ \\
% 0.4 & $66.26$ & $94.55$\\
% 0.5 &$65.38$ & $93.32$\\
% 0.6 &$65.55$ & $94.55$\\

% % FAR \cite{divya2022far} &  $16$ & $224\times224$ & Kinetics & $71.4$ \\
% % %Diff-FAR \cite{kothandaraman2022fourier} &  $8$ & $960\times540$ & Kinetics & $80.75$ \\
% % MITFAS &  $16$ & $224\times224$ & Kinetics & $78.6$ \\
% % \textbf{Ours} &  $16$ & $224\times224$ & Kinetics & \textbf{78.6} \\
% \bottomrule
% \end{tabular}
% }
% \caption{We evaluate the impact of the hyper-parameter $\alpha$ in the shifted leaky ReLu mapping using a subset of Diving48. Results shows that with $\alpha=0.3$, we get a better motion information distribution mapping which yields to a better accuracy.  }
% \vspace{-3pt}
% \label{tab:hyper}
% % \end{center}
% \end{table}

%% file: tables/dimension.tex
\begin{table}
\footnotesize
\centering
% \begin{center}
\resizebox{1.0\columnwidth}{!}{
\begin{tabular}{c c c}
\toprule
Size of Patch ($r\times r$) & \makecell{ UAV-Human \\ Top-1 Accuracy (\%)}  & \makecell{ Time cost\\ per frame (ms)}  \\
\midrule
$3 \times 3$ & $53.8$ & $5.4$ \\
$5 \times 5$ &  $54.6$ &  $6.8$\\
$7 \times 7$ & \textbf{55.0} & \textbf{9.2}\\
$11 \times 11$ & $52.3$ & $17.9$\\
$21 \times 21$ & $51.9$ & $67.8$\\
\bottomrule
\end{tabular}
}
\vspace{-7pt}
\caption{\textbf{Impact of the patch size.} A better tradeoff between accuracy and time cost is achieved with patch size $7 \times 7$.}
\vspace{-5pt}
\label{tab:dimension}
% \end{center}
\end{table}

%% file: tables/sthsth.tex
\begin{table}[t]
% \footnotesize
\centering
% \begin{center}
\resizebox{1.0\columnwidth}{!}{
\begin{tabular}{ccccccc}
\toprule
\multirow{2}{*}{Dataset} & \multicolumn{2}{c}{\textbf{SthSthV2}} & \multicolumn{2}{c}{\textbf{UCF101}} & \multicolumn{2}{c}{\textbf{HMDB51}} \\
                         & Top-1(\%)       & Top-5(\%)         & Top-1(\%)         & Top-5(\%)        & Top-1(\%)       & Top-5(\%)       \\
                         \midrule
Uniform                  &       57.5       &     84.3          &       94.3       &       99.3      &   64.6          &    88.1       \\
MG Sampler             &        59.2      &     85.2            &        94.6      &        99.2     &        64.9      &     87.9      \\
\textbf{Ours (Patch 7$\times$7)}  &       59.7 \textbf{(+0.5)}       &      85.2  \textbf{(+0)}       &         94.4 \textbf{(-0.2)}     &       99.6 \textbf{(+0.4)}   &         64.5 \textbf{(-0.4)}    &        87.2 \textbf{(-0.7)}      \\
\textbf{Ours (Patch 3$\times$3)}  &       60.3 \textbf{(+1.1)}       &      85.8  \textbf{(+0.6)}       &         95.1 \textbf{(+0.5)}     &        99.5 \textbf{(+0.3)}      &         65.5 \textbf{(+0.6)}     &        88.1 \textbf{(+0.2)}     \\
% \textbf{Ours} (Patch 3$\times$3) &       -        &      -        &              &             &              &           \\

\bottomrule
\end{tabular}
}
\vspace{-7pt}
\caption{\textbf{Results on SthSthV2, UCF101, HMDB51}. Given that the majority of videos in these datasets feature fixed cameras capturing close-range scenes with prominent human actors and minimal background changes, utilizing smaller patches (3$\times$3) can effectively capture finer details of the actions, leading to improved accuracy.}
\vspace{-10pt}
\label{tab:sth}
% \end{center}
\end{table}

%% file: 10_conclusion.tex
\section{Conclusion, Limitations and Future Work}
\label{sec:conclusion}
% We introduce patch mutual information (PMI) score to leverage the motion bias in the aerial videos. PMI score is used to represent the motion information between adjacent frames by measuring the similarity of frame patches via mutual information calculation. It quantifies how much discriminative motion information that is contained in one frame given another. PMI score provides an efficient yet robust way to distinguish the motion-salient frames from the aerial videos. Then, we propose an adaptive frame selection strategy based on shifted leaky ReLu and Cumulative Distribution Function. It first maps the PMI score to $[0,1]$ and enhances the motion bias between adjacent frames, making it easier to distinguish the motion-salient frames in videos. It divides the whole video into multiple segments where each segment convey the same amount of motion information. Adaptive sampling is then performed by randomly picking a representative frame from each segment so that all the key information about the motion are included for training. 

In this paper, we present a novel frame selection method for aerial video action recognition. We first introduce patch mutual information (PMI) score to represent the motion information between adjacent frames by measuring the similarity of frame patches via mutual information calculation. %It quantifies how much discriminative motion information that is contained in one frame given another. 
Then, we propose an adaptive frame selection strategy based on shifted Leaky ReLu and cumulative distribution function, which ensures that the sampled frames comprehensively cover all the essential segments with high motion salience. Our evaluations on multiple datasets illustrate the effectiveness of our method. Even though we improved the SOTA methods, the absolute accuracy on UAV-Human is not very high, and we will further explore the backbone architecture design to improve the accuracy.
%Limitations and Future Work: Our method is currently limited on aerial videos, and it would be useful to apply this technique to other domains. The absolute accuracy on UAV-Human is not very high, as compared to ground videos, and we need better methods to improve the accuracy.

%Even though we improved the SOTA methods, the absolute accuracy on UAV-Human is not very high, and we will further explore the backbone architecture design to improve the accuracy.

\textbf{Acknowledgement} This work was supported in part by ARO Grants W911NF2110026, W911NF2310046,  W911NF2310352  and Army Cooperative Agreement W911NF2120076

%% file: 12_appendix.tex
\appendix
\label{sec:appendix}

\section{Implementation Details}
\noindent  \textbf{Evaluation metrics}   We evaluate our method and other state-of-the-art methods using Top-1 and Top-5 accuracy scores, where the predictions are considered to be correct if the top 1 or top 5 highest probability answers match the actual label.

\noindent  \textbf{Implementation Details} All models in this paper are trained using NVIDIA GeForce 2080Ti GPUs and NVIDIA RTX A5000 GPUs. The initial learning rate is set at $0.1$ for training from scratch and $0.05$ for initializing with Kinetics pretrained weights. Stochastic Gradient Descent (SGD) is used as the optimizer with 0.0005 weight decay and 0.9 momentum. We use cosine/poly annealing for learning rate decay and multi-class cross entropy loss to constrain the final predictions. Unless further specified, the videos are decoded as a single clip and all the frames are randomly scaled and center cropped to the size $224 \times 224$ during training. During testing, we scale the shorter spatial side to $256$ and take 3 crops of $224 \times 224$ to cover the longer spatial axis. We average the scores for all individual predictions.

\section{Incorporate with different recognition backbone models}\label{sec:back}
We further demonstrate that our method could be used with different recognition backbone models to improve the accuracy. We compare the results using our PMI Sampler as well as uniform sampling and MG Sampler on 3 different backbones: SlowOnly-R50~\cite{feichtenhofer2019slowfast}, I3D~\cite{carreira2017quo} and X3D~\cite{feichtenhofer2020x3d}. Results on Table~\ref{tab:backbone} show that our method consistently outperforms other methods and brings accuracy improvement across different backbone models. 

\input{tables/backbone.tex}
\input{tables/denseclip.tex}
\section{Use in dense clip sampling}\label{sec:dense}
Our proposed PMI Sampler could also be used in dense clip sampling during training. We uniformly sample the videos into 10 clips and for each clip, we adaptively select 8 frames. The baseline method is to uniformly select those frames in each clip. We evaluate the performance of our proposed method along with the current state-of-the-art MG Sampler in dense clip sampling scenario. As shown in Table~\ref{tab:dense}, our method achieves a relative improvement over the baseline method by $2\%$ and $1.2\%$ over SOTA.

\section{Analysis}\label{sec:analysis}
In aerial videos, the human actor occupies less than $10\%$ resolutions and the rest pixels belong to the background. When the camera is moving, the overall background deviations are much larger than the actual motion changes. Therefore, pixel-wise RGB difference used in MG Sampler~\cite{Zhi2021MGSamplerAE} will be dominated by background noises and fails to map the motion distribution for both videos in Figure~\ref{fig:example}. However, our proposed patch mutual information is more robust because of the inherent advantage of mutual information. Mutual information measures the image similarity only by considering the overall pixel value distribution in the two images, see Eq~\ref{eq:miprob}. Thus, it is more robust to outliers and noises. However, it ignores spatial information between pixels and that is very important for action recognition. Patch mutual information avoids such issues by dividing the frames into patches and measuring the mutual information of small patches. In this way, the spatial information within the patches can be conserved.  Because of the robustness of PMI, we can further employ the shifted Leaky ReLu to make the motion-salient frames easier to distinguish. As shown in Figure~\ref{fig:example}, PMI Sampler ensures that the sampled frame comprehensively covers all the essential segments with high motion salience, so that key information about the somersault in Diving48 may not be missed. Also, PMI Sampler is robust to background noises. It can identify the motion static period even when the camera is shaking in UAV videos, see Figure~\ref{fig:example}, selecting more frames from the motion salient period and fewer frames from the motion static period.

\section{More Visualization Results}\label{sec:visualresults}

We generate more visualization results between our method and current state-of-the-art method, MG Sampler~\cite{zhi2021mgsampler}, on the three datasets: UAV-Human~\cite{li2021uav}, NEC-Drone~\cite{choi2020unsupervised} and Diving48~\cite{li2018resound} in Figure~\ref{fig:VS_D},\ref{fig:VS_U_L},\ref{fig:VS_U_S},\ref{fig:VS_F}.

As mentioned in Section~\ref{sec:method}, our proposed method quantifies the motion information contained in adjacent frames based on the similarity measure between corresponding frame patches. As for trimmed videos, human actors are performing scripted actions in the same scene and the backgrounds are always similar in the same video. Therefore, our patch similarity guided frame selection strategy is more robust to background noises. Moreover, since the backgrounds are similar, the unsimilarity is dominated by human actions, thus our method yields to better motion information representations. As shown in Figure~\ref{fig:VS_D},~\ref{fig:VS_U_S}, when the camera is moving and no salient motion exists, MG Sampler~\cite{zhi2021mgsampler} suffers from the pixel value changes corresponding to the backgrounds. However, our proposed PMI Sampler can accurately identify the motion static periods.

Our method also gives more accurate motion information distribution for aerial videos and makes it much easier to distinguish the motion salient frames, see Figure~\ref{fig:VS_U_L}. As mentioned in Section~\ref{sec:analysis}, this is attributed to our proposed patch mutual information score, which considers both the pixel distribution and spatial relationships inside the patches. Our method can select more frames from the motion salient periods and fewer frames from the motion static periods.

\section{Limitations}
Our proposed method may have two limitations. First, as shown in Figure~\ref{fig:VS_F}, when the motion is consistent and smooth (such as swing the racket, drink, rub hands) during the whole video, our method will perform just like uniform sampling. Second, in the case where the action label is highly associated with the gesture during motion static periods, our method may be less effective due to the reduced number of sampled frames during these periods. For instance, as depicted in the second video in Figure~\ref{fig:VS_F} with label "all clear", the action is primarily determined by the static gesture between frames 12 and 28. However, our approach samples fewer frames during this period, which might hinder the performance. To mitigate this issue, we can adjust the value of $\alpha$ in the shifted Leaky ReLu to achieve a smoother motion information distribution, enabling the selection of more frames during such periods. Nonetheless, further investigation is necessary to comprehensively address this concern.

\input{figs/VS_D}
\input{figs/VS_U_S}
\input{figs/VS_U_L}
\input{figs/VS_F}

%% file: tables/backbone.tex
\begin{table}[h]
\footnotesize
\centering
% \begin{center}
\resizebox{1.0\columnwidth}{!}{
\begin{tabular}{c c c c  c}
\toprule
Method & Frames & Backbone  & Top-1 Acc (\%) & Top-5 Acc (\%)   \\

\midrule
Uniform &  $8$& I3D~\cite{carreira2017quo}  & $59.2$ & $89.9$ \\
MG Sampler \cite{zhi2021mgsampler} &  $8$& I3D~\cite{carreira2017quo}  & $55.2 $ & $87.6$\\
\textbf{Ours} &  $8$& I3D~\cite{carreira2017quo}  & \textbf{61.8} & \textbf{91.7} \\

\midrule
Uniform &  $8$& SlowOnly-R50~\cite{feichtenhofer2019slowfast}  & $60.0$ & $90.3$ \\
MG Sampler \cite{zhi2021mgsampler} &  $8$& SlowOnly-R50~\cite{feichtenhofer2019slowfast}   & $57.1 $ & $88.4$\\
\textbf{Ours} &  $8$& SlowOnly-R50~\cite{feichtenhofer2019slowfast}   & \textbf{63.1} & \textbf{93.5} \\

\midrule
Uniform &  $16$& X3D~\cite{feichtenhofer2020x3d}   & $73.5$ & $95.1$ \\
MG Sampler \cite{zhi2021mgsampler} &  $16$& X3D~\cite{feichtenhofer2020x3d}   & $74.6 $ & $95.0$\\
\textbf{Ours} &  $16$& X3D~\cite{feichtenhofer2020x3d}   & \textbf{81.3} & \textbf{97.7} \\

\bottomrule
\end{tabular}
}
\caption{Evaluate our method with different recognition backbones on Diving48. PMI Sampler can be incorporated with any recognition backbone models to improve the accuracy. }
\vspace{-5pt}
\label{tab:backbone}
% \end{center}
\end{table}

%% file: tables/denseclip.tex
\begin{table}[ht]
\small
\centering
% \begin{center}
\resizebox{1.0\columnwidth}{!}{
\begin{tabular}{c c c c c}
\toprule
Method & Frames & Backbone  & Top-1 (\%) &  Top-5 (\%)   \\
\midrule
Uniform &  $8$ &X3D-M~\cite{feichtenhofer2020x3d}& $74.0$ & $95.3$\\
MG Sampler \cite{zhi2021mgsampler} & $8$&X3D-M~\cite{feichtenhofer2020x3d} & $74.6 $ & $95.9$\\
\textbf{Ours} &  $8$ &X3D-M~\cite{feichtenhofer2020x3d}& \textbf{75.5} & \textbf{96.0} \\

% FAR \cite{divya2022far} &  $16$ & $224\times224$ & Kinetics & $71.4$ \\
% %Diff-FAR \cite{kothandaraman2022fourier} &  $8$ & $960\times540$ & Kinetics & $80.75$ \\
% MITFAS &  $16$ & $224\times224$ & Kinetics & $78.6$ \\
% \textbf{Ours} &  $16$ & $224\times224$ & Kinetics & \textbf{78.6} \\
\bottomrule
\end{tabular}
}
\caption{Our propposed PMI Sampler can be used in dense clip sampling for improved accuracy. We demonstrate an relative improvement in top-1 accuracy over baseline method by $2\%$ and $1.2\%$ over SOTA.}
\vspace{-8pt}
\label{tab:dense}
% \end{center}
\end{table}

%% file: figs/VS_D.tex
\begin{figure*}[tp]
    \centering
    \includegraphics[width=\textwidth]{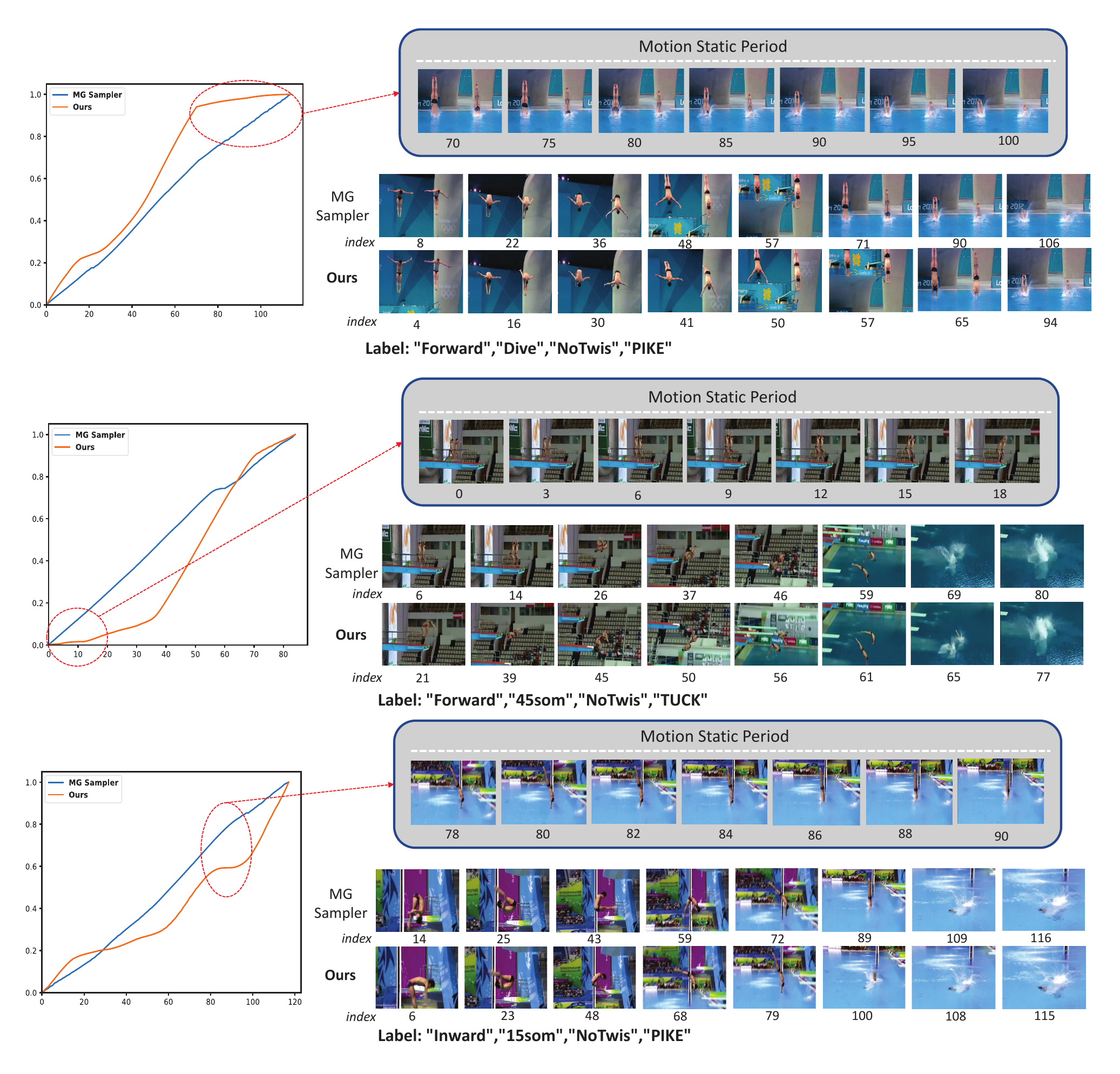}
    % \includesvg[width=\linewidth]{figs/ReLu.svg}
    \vspace{-5mm}
    \caption{Comparison between our method and MG Sampler~\cite{zhi2021mgsampler} on typical videos from Diving48~\cite{li2018resound}. As shown above, MG Sampler fails to measure the motion information between frames and cannot reflect the motion distribution of the video because of the background changes caused by the camera moving. However, our method is more robust to background noises and could accurately identify the motion static periods in the start, middle and end of the videos.} 
    % Our method select more frames from motion salient period and less frames from static periods.}

    \label{fig:VS_D}
    \vspace{-3mm}
\end{figure*}

%% file: figs/VS_U_S.tex
\begin{figure*}[tp]
    \centering
    \includegraphics[width=\textwidth]{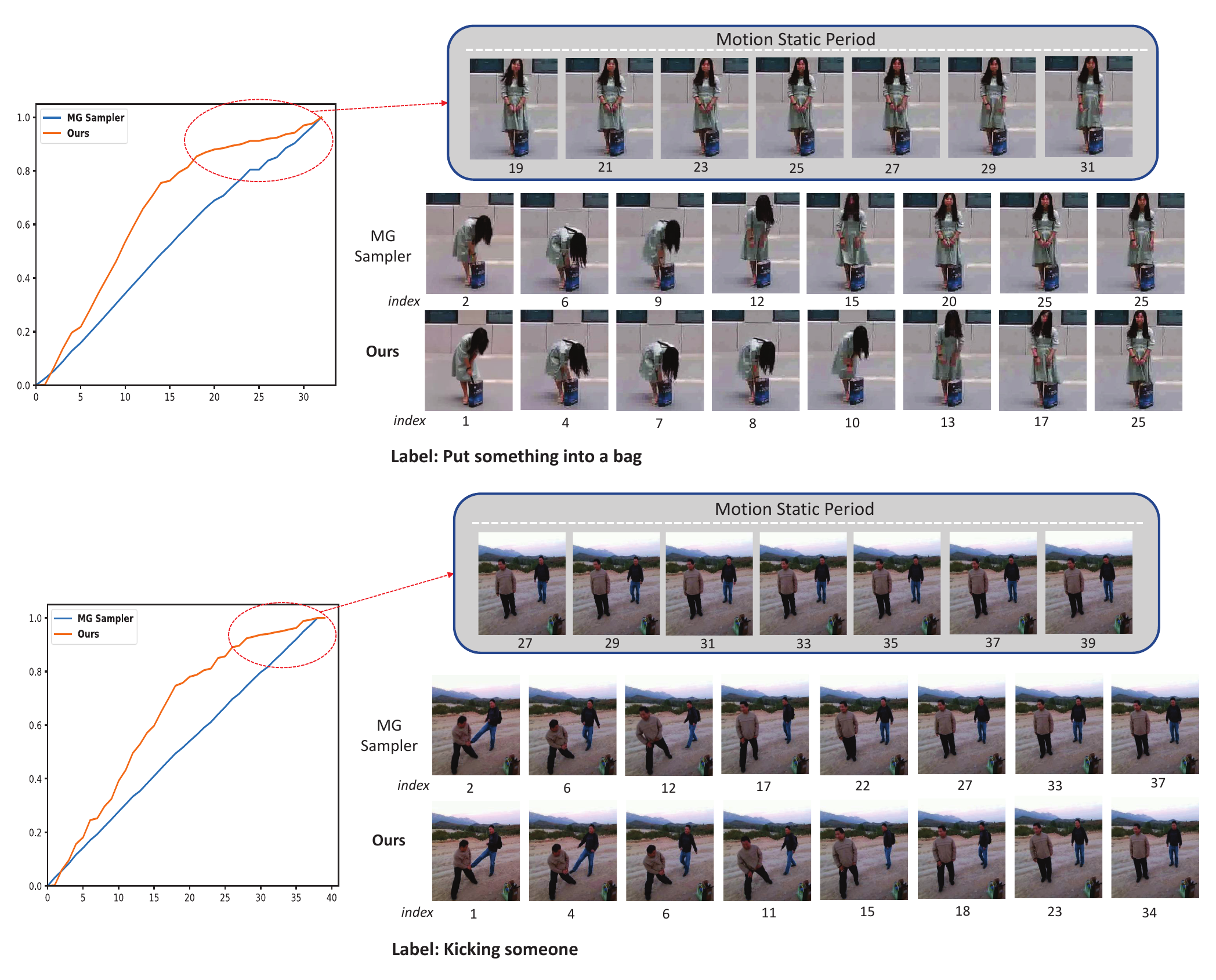}
    % \includesvg[width=\linewidth]{figs/ReLu.svg}
    \vspace{-5mm}
    \caption{Comparisons between our method and MG Sampler~\cite{zhi2021mgsampler} on typical videos from UAV-Human~\cite{li2021uav} and NEC-Drone~\cite{choi2020unsupervised}. Compare to Diving48~\cite{li2018resound}, UAV videos are more shaky and most pixels are corresponding to backgrounds(frames in the figure are cropped for better visualization). Therefore, they contain more background noises. MG Sampler fails to handle such challenges from UAV videos. However, due to the robustness of our proposed patch mutual information, our method could accurately distinguish the motion static period. } 
    % Our method select more frames from motion salient period and less frames from static periods.}

    \label{fig:VS_U_S}
    \vspace{-3mm}
\end{figure*}

%% file: figs/VS_U_L.tex
\begin{figure*}[tp]
    \centering
    \includegraphics[width=\textwidth]{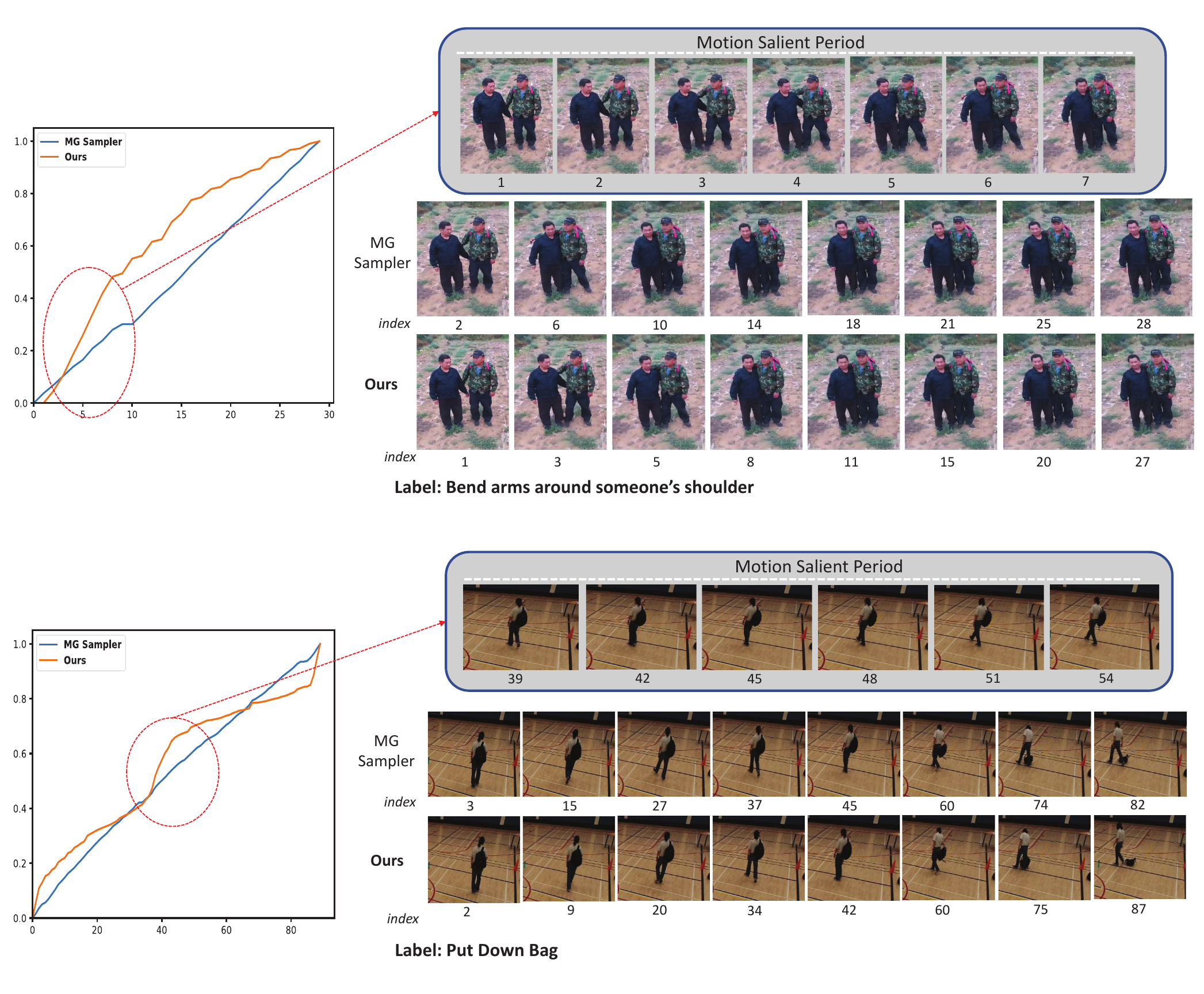}
    % \includesvg[width=\linewidth]{figs/ReLu.svg}
    \vspace{-5mm}
    \caption{More comparisons between our method and MG Sampler~\cite{zhi2021mgsampler} on typical videos from UAV-Human~\cite{li2021uav} and NEC-Drone~\cite{choi2020unsupervised}. Our method makes it easier to distinguish the motion salient frames. Our method selects more frames from the motion salient periods and less frames from motion static period, so that sampled frames contain more useful motion information. } 
    % Our method select more frames from motion salient period and less frames from static periods.}

    \label{fig:VS_U_L}
    \vspace{-3mm}
\end{figure*}

%% file: figs/VS_F.tex
\begin{figure*}[tp]
    \centering
    \includegraphics[width=\textwidth]{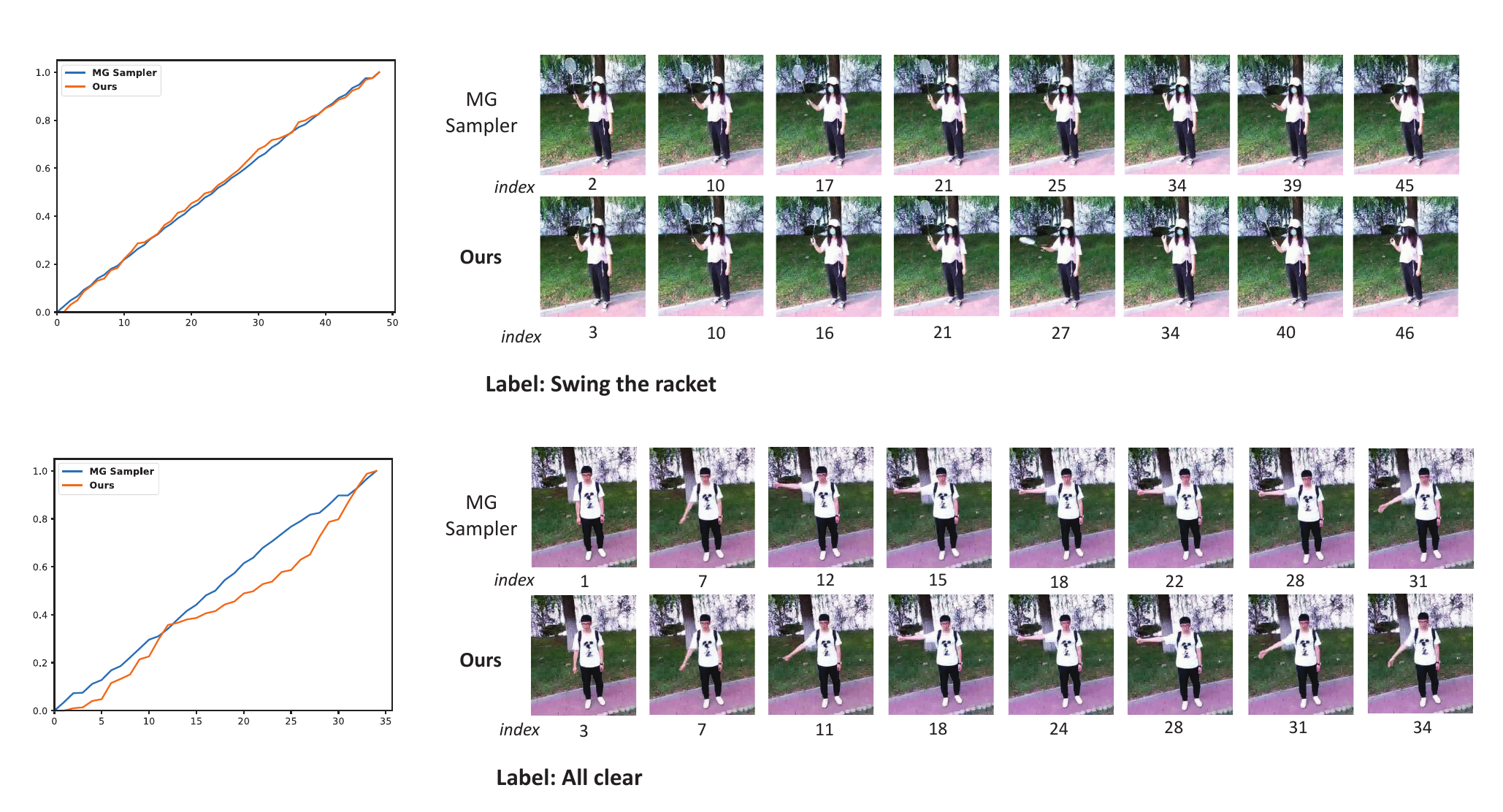}
    % \includesvg[width=\linewidth]{figs/ReLu.svg}
    \vspace{-5mm}
    \caption{There are two limitations of our method. First, as shown in the first video (with label: swing the racket), when motion is consistent and smooth during the whole video, our method will perform just like uniform sampling. Second, in instances where the action label is highly associated with the gesture during motion static periods, our method may be less effective due to the reduced number of sampled frames during these periods. As shown in the second video above (with label: all clear), the action is primarily determined by the static gesture between frame 12 to frame 28. However, our method samples fewer frames from such period. To mitigate this issue, we can adjust the value of $\alpha$ in the shifted Leaky ReLu to achieve a smoother motion information distribution, enabling the selection of more frames during such periods. Nonetheless, further investigation is necessary to comprehensively address this concern. } 
    % Our method select more frames from motion salient period and less frames from static periods.}

    \label{fig:VS_F}
    \vspace{-3mm}
\end{figure*}